\documentclass[sigconf,natbib=true,anonymous=false]{acmart}
\usepackage{amsmath,nccmath,tabularx} 
\usepackage{algorithm}
\usepackage{algorithmic}
\usepackage{color}
\usepackage{graphicx}
\graphicspath{{images/}}
\usepackage{wrapfig,lipsum}
\usepackage{tabularx}
\usepackage{graphicx}
\usepackage{lscape}
\usepackage{subcaption}
\usepackage{latexsym}
\usepackage{pifont}
\usepackage{multirow}

\usepackage{enumitem}
\setlist{leftmargin=5mm}
\usepackage[T1]{fontenc}
\usepackage[utf8]{inputenc}
\usepackage[font=small,labelfont=bf]{caption}

\AtBeginDocument{%
  \providecommand\BibTeX{{%
    \normalfont B\kern-0.5em{\scshape i\kern-0.25em b}\kern-0.8em\TeX}}}

\copyrightyear{2023} 
\acmYear{2023} 
\setcopyright{acmlicensed}\acmConference[SIGIR-AP '23]{Proceedings of the 1st International ACM SIGIR Conference on Information Retrieval in the Asia Pacific}{November 26--29, 2023}{Beijing, China}
\acmBooktitle{Proceedings of the 1st International ACM SIGIR Conference on Information Retrieval in the Asia Pacific (SIGIR-AP'23, November 26--29, 2023, Beijing, China}
\acmPrice{15.00}
\acmDOI{xxxxxx}
\acmISBN{978-1-4503-8037-9/21/07 Version submitted to SIGIR-AP on 4 July 2023.}

\begin{document}

\title{ChatGPT Hallucinates when Attributing Answers}



\author{Guido Zuccon}
\affiliation{%
	\institution{The University of Queensland}
	\city{Brisbane}
	\state{QLD}
	\country{Australia}}
\email{g.zuccon@uq.edu.au}

\author{Bevan Koopman}
\affiliation{%
	\institution{CSIRO \& The University of Queensland}
	\city{Brisbane}
	\state{QLD}
	\country{Australia}}
\email{bevan.koopman@csiro.au}

\author{Razia Shaik}
\affiliation{%
	\institution{The University of Queensland}
	\city{Brisbane}
	\state{QLD}
	\country{Australia}}
\email{r.shaik@uq.edu.au}

\begin{abstract}
Can ChatGPT provide evidence to support its answers? Does the evidence it suggests actually exist and does it really support its answer? We investigate these questions using a collection of domain-specific knowledge-based questions, specifically prompting ChatGPT to provide both an answer and supporting evidence in the form of references to external sources. We also investigate how different prompts impact answers and evidence. 

We find that ChatGPT provides correct or partially correct answers in about half of the cases ($\approx50.6\%$ of the times), but its suggested references only exist 14\% of the times. We further provide insights on the generated references that reveal common traits among the references that ChatGPT generates, and show how even if a reference provided by the model does exist, this reference often does not support the claims ChatGPT attributes to it.

Our findings are important because (1) they are the first systematic analysis of the references created by ChatGPT in its answers; (2) they suggest that the model may leverage good quality information in producing correct answers, but is unable to attribute real evidence to support its answers. 
Prompts, raw result files and manual analysis are made publicly available at \url{https://www.github.com/anonymized}.


\end{abstract}


\maketitle

\section{Introduction}

Large Language Models (LLMs) such as ChatGPT are increasingly been used by people for information seeking tasks. They provide a convenient way to access information, as questions can be posed in natural language and answers are provided in a fluent, summarised and often easy to understand form. This allows for more rapid and less laborious access to information, compared to other methods for information seeking, such as search engines. 

While LLMs have shown impressive effectiveness in answering questions that require factual responses~\cite{rae2021scaling}, they have been shown to often provide incorrect or not actionable answers when dealing with commonsense knowledge, logical reasoning and common misconceptions~\cite{rae2021scaling,lin2021truthfulqa,zuccon2023dr}.

\begin{figure}[t]
	\includegraphics[width=1\columnwidth]{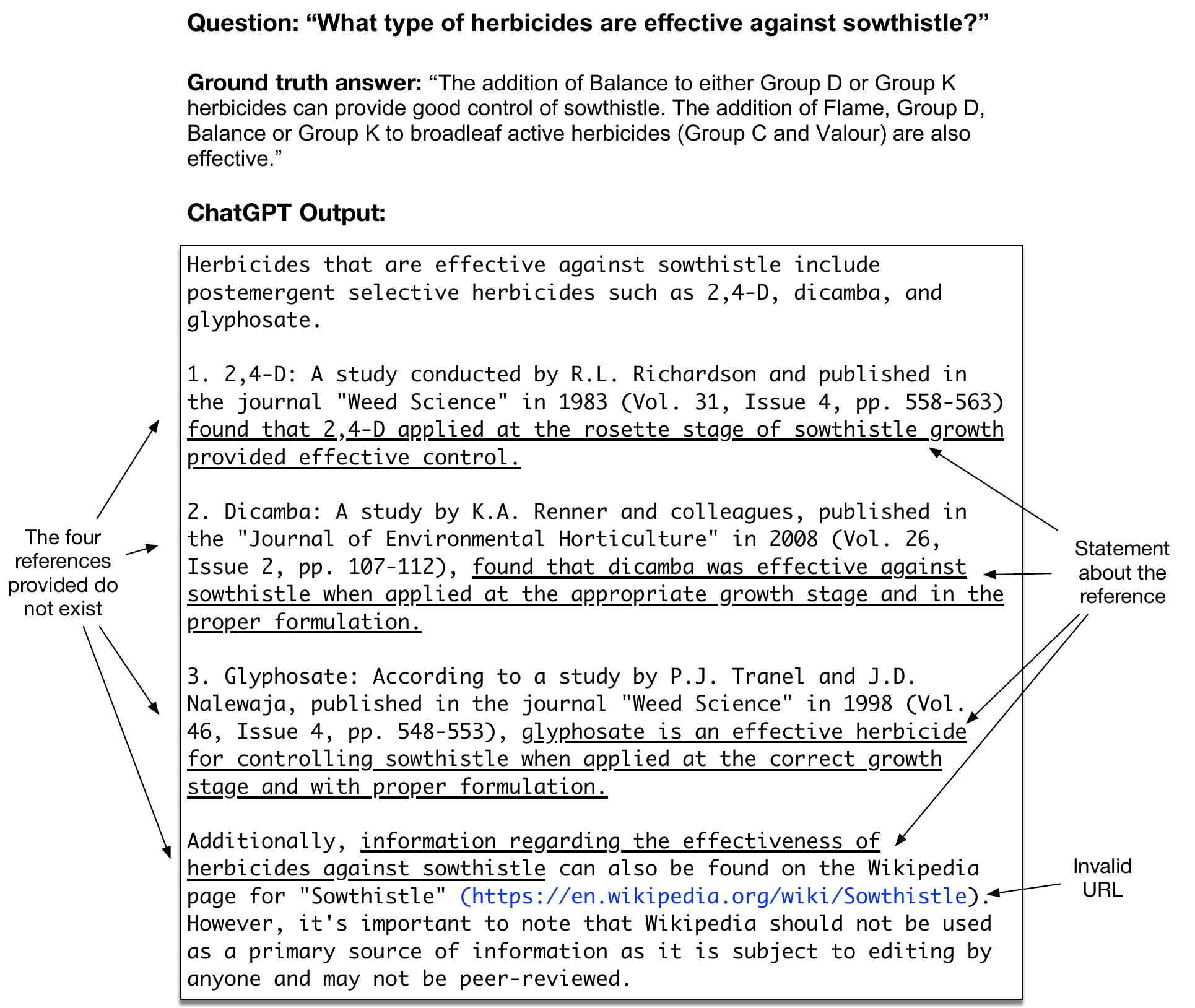}
	\caption{Sample session with ChatGPT. Four references are provided as evidence. While they appear like real reference, they do not actually exist. In addition, the statement about the Wikipedia reference is not supported by the actual content of the article. \label{fig-sample}}
\end{figure}

A key direction that has been suggested to support information seeking tasks and address answer errors by empowering users to verify the answers generated from LLMs, is the ability of a LLM to attribute their answer~\cite{metzler2021rethinking,bohnet2022attributed,rashkin2023measuring,menick2022teaching,thoppilan2022lamda,gao2022attributed}. Attribution is the ability to generate evidence, in the form of a reference or citation, that supports claims the LLM makes in its answer. Figure~\ref{fig-sample} provides an example of a question and an answer from ChatGPT with four references provided by the model to attribute the answer.

Recent research has investigated techniques for attribution~\cite{metzler2021rethinking,bohnet2022attributed,rashkin2023measuring,menick2022teaching,thoppilan2022lamda,gao2022attributed}; but the validity of the attributions produced by the popular ChatGPT model, which at the time of writing boasts a user base in excess of 100 million people, has received little attention. This is an important aspect of ChatGPT to investigate, as anecdotal reports day the model  hallucinates attributions; i.e., attributions that appear realistic, but that do not exist in reality. An example of this is shown in Figure~\ref{fig-sample}, where none of the four references provided, included a Wikipedia page\footnote{This is interesting to observe because ChatGPT's creators, the OpenAI company, have stated that the model was trained on Wikipedia pages.}, exist. 

In this paper, we aim to provide a thorough analysis of attribution generated by ChatGPT. To this aim, we perform a qualitative analysis of ChatGPT responses and generated references for a large dataset of domain-specific knowledge-based questions, and we answer the following research questions:



\begin{enumerate}[label=\bf RQ\arabic*, align=left, leftmargin=*]
  \item Can ChatGPT provide supporting evidence (in the form of references) to its answers?
  \item Does the supporting evidence/references actually exist (i.e., not hallucinated)?
  \item Do the statements provided by ChatGPT about the evidence actually align with what the reference says?
\end{enumerate}

\section{Related Work}

The recent progress in instruction-based large language models (LLMs), for instance, ChatGPT, has demonstrated their proficiency in adhering to user guidelines to successfully accomplish tasks~\cite{gozalo2023chatgpt, guo2023close,wang2023can, sallam2023chatgpt}. These models, typically possessing tens of billions of parameters, are pre-trained on a wide range of substantial text data. This allows them to create pertinent and coherent responses on a diverse selection of topics~\cite{gozalo2023chatgpt}.
Various studies have evaluated ChatGPT's performance on a number of different downstream tasks, consistently noticing an improved efficiency in task resolution, such as question answering~\cite{tan2023evaluation, omar2023chatgpt}, and ranking~\cite{sun2023chatgpt, ji2023exploring}. 

LLMs are increasingly been used for information seeking tasks, ranging from straightforward question-answering situations~\cite{bohnet2022attributed}.
However, LLMs are fronted with a number of open challenges~\cite{weidinger2022taxonomy}, including hallucinations and correct attribution of answers, which currently limit how much people should rely on these tools for information seeking. 

Hallucinations in LLMs like ChatGPT refer to the generation of factually incorrect or entirely made-up information~\cite{ji2023survey}. These models are trained to generate text based on patterns they have learned from the input data, often through the next token prediction task, and while they are adept at creating coherent and plausible-sounding responses, they sometimes produce outputs that are not grounded in reality. Hallucinations have been reported to occur more often when common sense and logical reasoning is required~\cite{rae2021scaling}, or when dealing with common misconceptions~\cite{lin2021truthfulqa,zuccon2023dr}, while LLMs appear to perform more robustly on tasks requiring factual responses~\cite{rae2021scaling} -- though our results will show this not to be necessarily the case within the specific domain of the question set we consider. 
One of the key reasons for this is that these models do not have a true understanding of the world or access to real-time information. While the presence of frequent, incorrect statements in the training data may let the model learn to reproduce similar incorrect statements, even with a diverse and high-quality training dataset, LLMs can still hallucinate information due to the way they handle uncertainty. Given a prompt that is ambiguous or open-ended, the model may generate text that seems reasonable but is completely fabricated. The reliance on retrieve-then-generate pipelines, where a generation occurs from evidence retrieved through an initial round of search, and the use of attribution have been poised to be possible mitigation strategies~\cite{weidinger2022taxonomy}. 

Attribution refers to the ability of a LLM to provide evidence (in the form of a snippet, citation or reference) that supports the answer, or part of, that it generates~\cite{metzler2021rethinking,bohnet2022attributed,rashkin2023measuring,menick2022teaching,thoppilan2022lamda,gao2022attributed,gao2023enabling}. The availability of such evidence would enable the verifiability of the answer~\cite{gao2023enabling} -- i.e., a user could follow the reference made by the LLM to verify whether the cited source supports the provided answer. It has also been suggested that attribution could improve the factual correctness of the answers~\cite{gao2023enabling}; however in our experiments we show that factual correctness is relatively low despite answers having attributions, at least for the commonly used ChatGPT LLM. We notice that while the quality and correctness of citations and references produced by LLMs have been somewhat investigated in the context of specific techniques aimed at offering attribution capabilities~\cite{menick2022teaching,gao2023enabling,metzler2021rethinking,bohnet2022attributed}, the evaluation of attribution references produced by the popular ChatGPT has been limited~\cite{gravel2023learning,hueber2023quality}. In this paper we aim to address this gap by providing a qualitative analysis of ChatGPT's attribution capabilities and the references it generates using a large datasets of domain-specific knowledge-based questions.


\section{Methodology}

To answer our research questions regarding the quality of the attributions made by ChatGPT in its answers, we prompted the model with domain-specific knowledge-based questions.  
As questions to submit to ChatGPT, we used the topics from the Ag-valuate collection~\cite{koopman2023agask}, a test collection for both passage and document retrieval in the Agriculture domain. The topics consist of natural language questions created by agricultural scientists and crop growers, and the collection contains a total of 160 topics\footnote{We only used the training portion of the topics for our experiments.}. Along with questions, the collection also provides a topic-creator authored answer, which forms the ground truth, and sparse relevance judgements over a dataset of 9M+ passages extracted from specialised agricultural websites and scientific publications. An example question and ground truth answer is shown in Figure~\ref{fig-sample}. We selected this collection for the experiment because (i) it was public and readily available, (ii) it provided ground truth answers, along with evidence to sources for the answers, thus ensuring that such source do exist, (iii) we had access to a domain expert that could interpret the answers and the evidence provided in support.

\begin{figure}
	\ttfamily  \raggedright
	Consider my question: {[question\_text]} \\[4pt]
	Now, provide evidence for my question, for example research articles, articles from specialised magazines, Wikipedia pages. If referencing a research article or magazine, provide the name of authors, title of the artitle, publication venue (including volume, issue, page numbers), year. If referencing Wikipedia or a web page, provide the page name and the URL. 
	\caption{GPTChat prompt format.}
	\label{fig-prompt}
\end{figure}

Questions were issued to the online version of ChatGPT (in the GPT3.5 version), and answers collected. In issuing questions, we embedded them into the simple prompt of Figure~\ref{fig-prompt}, which instructs the model to answer the question and provide evidence for the answer.

ChatGPT's answers were examined by the first two authors of the paper along with a research assistant, all computer scientist. This group of assessors were responsible to identify whether the answer contained references, and if it did, then they went on to locate the evidence, if it existed. Subsequently, a fourth annotator, the third author of this paper, analysed once again the answers. This annotator is an agricultural scientist in a leading university in Ag-Tech, and is expert in the crops -- the topic area the questions in the Ag-valuate collection focus on.
This annotator validated the previous annotations, correcting for errors. In addition, she validated the answers of ChatGPT for correctness and provided further comments regarding the model's answers. Finally, she also examined the references that were included in ChatGPT's answers and that we successfully located. For each of these, she assessed whether the reference contained the claims ChatGPT made with its regards. 

Annotations were provided with respect to the following annotation schema:

%
%
\begin{enumerate}
  \item Was the answer provided correct? Specifically, does the answer provided by ChatGPT align with the ground truth answer? Options: ``Yes, fully'', ``Only partially'', ``No (or no to a large extent)''.
  
  \item Are references provided in the answer? This was a binary question (yes/no), if yes, then the annotators had to select whether references were (i) ``Academic publication (journal, conference)'', (ii) ``Professional magazine/online publication'', (iii) ``Wikipedia page'', (iv) ``Other specialised website''.
  
  \item For each reference, does it actually exist? For a journal reference, does the paper with that title exist? Does the suggested Wikipedia article exist? Options: ``Yes'', ``No''; in addition, annotators were to add a comment regarding the source to reflect whether the journal existed (including issue etc.), but not the article, or the website existed, but not the specific page.
  
  \item Is a URL provided with the references that actually points to mentioned source? Options: ``Yes'', ``No''.
  
  \item Does the statement about the reference in ChatGPT's answer align with the actual reference content? Options: ``Yes, fully'', ``Only partially'', ``No (or no to a large extent)''.
\end{enumerate}

In addition, the annotators could add comments regarding the correctness of the answers and of the references, or any other observation worth noting. 

\section{Results}

\subsection{Answer Correctness}

Before diving into the analysis of the results for our research questions, we investigate the correctness of the answers provided by ChatGPT. Recall that the answers were assessed for correctness by the third author of this paper, an Ag-Tech expert. 

We found that a large portion of the provided answers were incorrect (49.4\%), with only 13.1\% being either fully correct and and 37.5\% partially correct.

Some of the incorrect answers were not just wrong, they were misleading and disastrous too. For example, for the question ``What type of herbicides are effective against sowthistle?'', ChatGPT lists ``2,4-D'', ``Dicamba'' and ``glyphosate''. However, sowthistle is resistant to ``2,4-D'' and ``glyphosate'', and in parts of the world, e.g. in part of Australia, sowthistle is also resistant to ``Dicamba''. This meant that growers that  to followed ChatGPT's suggestion, would spend a considerable amount of money to purchase and apply these herbicides, without obtaining the intended outcome. These herbicides also have disadvantages. For example, glyphosate products are harmful to animals if they touch or eat plants treated with it; some studies have also suggested glyphosate may be linked to cancer (and is classified as a probable human carcinogen), though others have suggested there is no link~\cite{myers2016concerns}.

Other answers were wrong in the context of the question, but their content was not necessarily incorrect. For example, for the question ``What is the best treatment for soybean to improve its digestibility?'', the model provides an open ended answer, where there is no clear identification of a best treatment, nor a comparison between the treatment options it identifies. 

\subsection{RQ1: Is there Supporting Evidence?}

Our first research question aimed to assess whether ChatGPT produced evidence, in the form of references to publications or websites, to support the answers it produced. Despite the model being explicitly told that references should be contained in the answer, we found that 14 out of the 160 answers (9\%) did not contain references. 
For the answers that included references (the remaining 91\%), there were on average 3.08 references per answer.

\begin{figure}[t]
  \includegraphics[width=0.85\columnwidth]{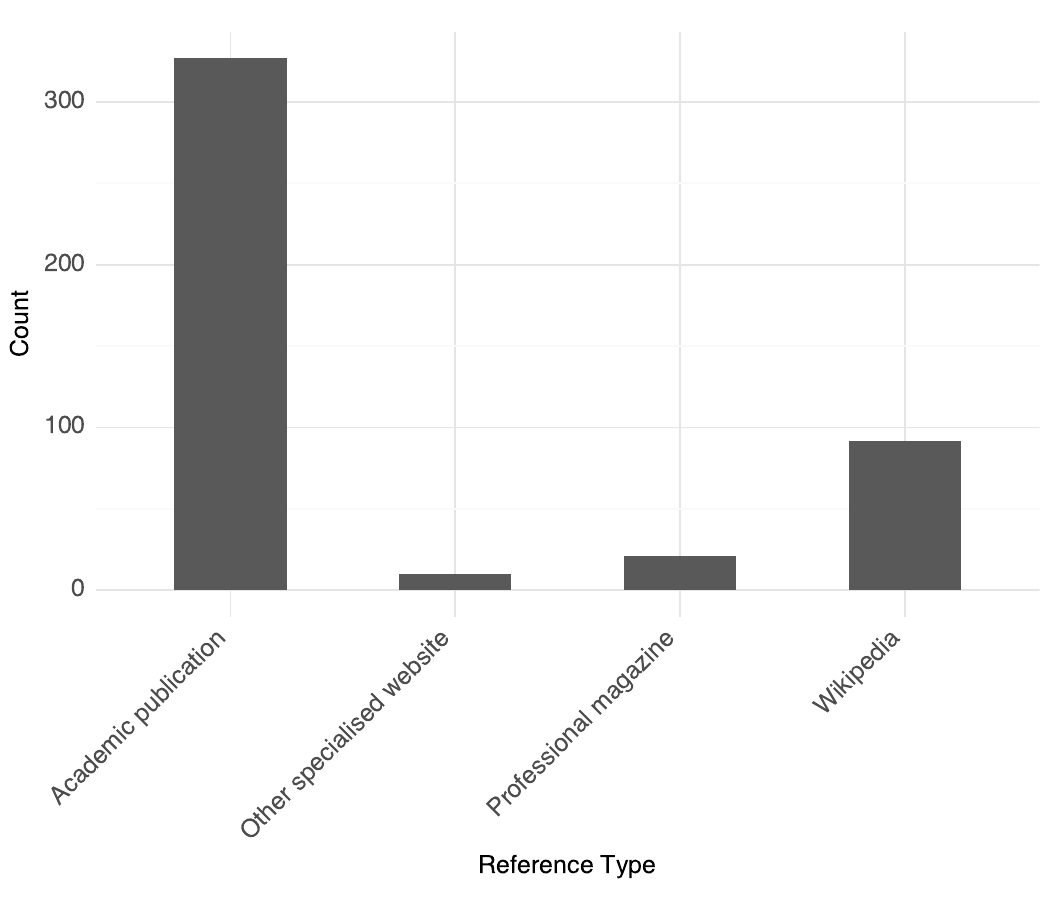}
  \caption{Distribution of references across four reference categories. Most references are academic publications. \label{fig-ref-type}}
\end{figure}

Figure~\ref{fig-ref-type} shows the distribution of reference types for the 450 references that were provided. Most references generated by ChatGPT are academic publications, followed by Wikipedia pages. 

When examining the answers that did not contain references, we identified some specific patterns. 
Often, this occurred when the model stated it did not know how to answer the questions. 
Other times, when the model did provide an actionable answer but no evidence in the form of a reference, it did  suggest how to find references. In these cases, the model may suggest how the user could attempt to identify potential evidence themselves. At times, it did so giving high level suggestions, e.g., for question ``When does awnless barnyard grass germinate?'', it suggested ``you may want to search for more general information on the germination of warm-season annual grass species, which can provide insight into the factors that influence the timing of germination for awnless barnyard grass.''. Other times, it provided more specific information, including data repositories to search.
For example, for the question ``Are deep phosphorus applications economic in Central Queensland?'', the answer of the model explicitly suggested to search agricultural journals or online databases such as Google Scholar, JSTOR, or the Agriculture and Applied Economics Association. Other times the model mentioned studies or publications but in a way that it was too generic to identify a single source that we could check. For example, for the question ``How much water can I lose if i let my cover crop grow for too long?'', ChatGPT mentioned ``According to a publication by the USDA Natural Resources Conservation Service'', ``A study by the University of California Division of Agriculture and Natural Resources'',  and``a publication by the National Center for Appropriate Technology''. However, it did not provide specific details such as authors, venue or URL that could have helped identifying these references. 
There were cases however, when the model while providing an answer, it simply did not provide any reference or help to locate one. In the few cases in which this happened, ChatGPT would apologise for not been able to find specific articles or Wikipedia pages --- though it then resorted to claim there was evidence (not better specified) to answer the question.

\subsection{RQ2: Does the Evidence Exist?}

Next, we investigate whether the evidence provided by ChatGPT as part of a reference actually exists. The model generated in total 450 references for the 160 questions we submitted. Of these references, 385 (86\%) did not exist, demonstrating the large extent of hallucinations ChatGPT produces with respect to the attribution of its answers.

When we analysed the references generated by ChatGPT and that we could not locate, we often were able to locate the journals mentioned by the model. These journals were most often high quality journals in the Agricultural Science and Tech space. 
In that journal, we could find the volume and issue provided in the answer, which corresponded to the publication year the model mentioned. Page ranges also matched as being in these volumes/issues; however, they were incorrect as often did not match the exact start/end pages of articles, and instead referred to incorrect article boundaries. Nonetheless, articles with the titles provided by ChatGPT did not exist in these journals, nor at all existed when we searched for them via Google or on Google Scholar. We also noted that often the authors ChatGPT provided in the references generated were actual researchers --- but not necessarily in the Ag Tech space; e.g., some were prominent medical scientists. There were cases in which an article that was provided as a reference was actually found by title, but with different authors, volume, issue and year data. 

We then analysed the references that did exist: 14\% of the total references provided by ChatGPT. Of these, the large majority (85\%) were references to Wikipedia pages, with the remaining approximately distributed among the remaining three types of references. (Note, we checked if a page with that title existed, and did not check the URL; we discuss the existence of URL separately below). It is not surprising that generated references to Wikipedia pages were found to exist: (1) ChatGPT was trained on a corpus containing  Wikipedia pages; (2) the format of the title of Wikipedia pages has been standardised with explicit style guides\footnote{e.g., \url{https://en.wikipedia.org/wiki/Wikipedia:Article_titles}.}, and thus potentially it is fairly easy to guess the title of a possible Wikipedia page for a topic. This becomes evident when comparing the titles of Wikipedia pages provided by ChatGPT that do exist with those that do not exist.

Sometimes references contained an URL. This was either because the reference was a web page (92 were Wikipedia pages and 12 were other type of web pages), or because the URL was the DOI associated with a publication. Note that in the analysis above that considered whether an evidence existed, we did not consider whether the URL itself existed\footnote{In the case of the reference being a Wikipedia page or webpage, we used the title of the page to check on the mentioned website if that page existed.}. We then analyse the URLs provided by ChatGPT next.
First, we note that when giving a Wikipedia page as reference, ChatGPT would seldom also provide an URL to the page.
For the URLs that were produced, we did observe that they appeared to be realistic. In particular, Wikipedia pages had URLs that matched the stylistic guidelines used by Wikipedia. URLs that referred to DOIs also appear to follow the typical structure for DOIs, and URLs that corresponded to relevant agencies in Ag-Tech, like the United States Department of Agriculture (USDA) or the Queensland Department of Agriculture And Fisheries (DAF) were also mostly following the format of URLs these entities commonly use -- with especially the website domain and the first level URL path being correct. However, the large majority of these URLs did not exist: only 34 existed. Of the URLs that existed, most were Wikipedia pages. We remind the reader that Wikipedia was used in the training of ChatGPT, and that Wikipedia pages have a rather straightforward URL structure: if a Wikipedia page about a topic/key-term \texttt{X} exist, then it is easy to guess its URL to be \url{https://en.wikipedia.org/wiki/X}.

The fact that the reference mentions provided by ChatGPT look realistic, both in cases on the bibliography data of scientific articles, or the URL format of web pages, but these reference more than often do not exist, adds to the challenges posed by hallucinations. A user that examines ChatGPT answers, may believe the answers to be correct because references are provided that look like those one would expect: from titles being topical and likely similar to those one would use in a scientific article, to cited journal being among the top in the field, and websites and Wikipedia pages looking topical in terms of provided title, and credible in terms of provided source. It is only if the user were to perform some due diligence by attempting to locate the cited references that they would not be able to locate these, and thus possibly question whether ChatGPT's answers are correct.

\subsection{RQ3: Do References Support the Claims Made by ChatGPT?}

\begin{figure}[t]
	\includegraphics[width=1\columnwidth]{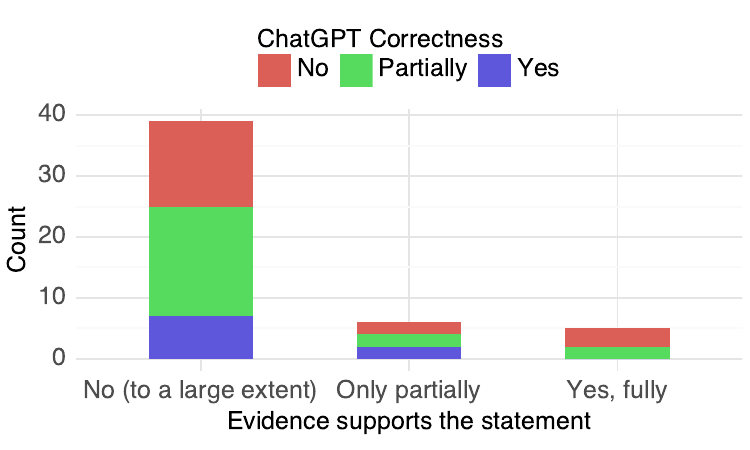}
	\caption{Breakdown of supporting evidences by level of support the reference provides and the correctness of ChatGPT's answer. \label{fig-breakdown-refs}}
\end{figure}

Next, we investigate whether the references provided by ChatGPT as evidence to support the answers it provides are indeed aligning with the claims made by the model.
For this analysis, we consider only references provided by ChatGPT that existed online. There were a total of 50 references that we could identify as existing. Figure~\ref{fig-breakdown-refs} provides a breakdown of the extent to which these references supported the claims advanced by ChatGPT, along with the correctness of the corresponding answers.

We analyse each of the three cases above separately to identify common patterns.  We start by analysing those references that fully supported the answer. For four out the five questions for which this happened, the reference that provided supporting evidence for the ChatGPT's claim was a Wikipedia page: this is not surprising as this page would have likely been seen in the model's training data. For the remaining question, the reference was a scientific journal article --- and this was one of the few questions for which ChatGPT provided a journal article that actually existed. However, the expert Ag Tech annotator noted that this article did provide very high quality evidence to answer the question (``What impact does P have on soil organic carbon levels following the introduction of a pasture?''), but the ChatGPT answer's itself did not use this evidence. This is to the extent that the answer provided by ChatGPT was incorrect, with no or little connection to the actual question, failing to touch upon phosphorus (P). In fact, though for all these five questions the evidence existed and it fully supported ChatGPT's claims, the answers provided by the model were labelled as incorrect (3 out of 5) or only partially correct (the remaining 2).

We then move to analyse the references that were considered by the annotator as only partially supporting ChatGPT's answer: there were just six of such cases. Five out of six references were Wikipedia pages, which we were able to locate, and the remaining reference was a scientific journal article, which we were able to locate using the article title, but noted ChatGPT attributed it to the wrong authors and year of publication. For all these cases the annotator highlighted how the evidence provided only partially backed up ChatGPT's answer; e.g., for the question ``Does high pH reduce productivity of grain crops?'' and the answer ``Yes, high pH levels can indeed reduce the productivity of grain crops.'', the Wikipedia page used by ChatGPT as reference provided a detailed explanation of soil pH, but did not directly and specifically provide evidence to support the answer. For one specific question (``What can be done to control powdery mildew in mungbeans?''), the annotator highlighted how the evidence provided by ChatGPT, a Wikipedia page about powdery mildew, did partially support the model's answer, but it also provided far too many more control options, e.g., genetic resistance, that ChatGPT instead failed to summarise in its reply. In other words, ChatGPT was not able to exploit the knowledge available in the references it provided to support its answers. In terms of answer correctness, for answers with references that only partially supported ChatGPT's claims, we found that correctness level was evenly spread across the three annotation levels.

Finally, we analyse the references that provided no support to ChatGPT's answers: there were 39 of these cases (78\%). We found that 87\% of these references were Wikipedia pages. Of the remaining, four were professional magazines or handbooks (specialised publications), two were scientific articles. Three of the four specialised publications were from government agency (the United States' USDA and the Australia's DAF and their federal Bureau of Agricultural and Resource Economics and Sciences). The remaining publication was an handbook, for which the reference metadata was only partially correct (year and authors  incorrect) and its content was regarded as completely out of context for the question by the expert Ag-Tech annotator. The two scientific articles also had issues with respect to incorrect metadata (authors, years), despite us being able to find the articles when searching by title; nevertheless their content had no bearing with the question or the answer. 
We also found that most answers associated to references that did not support claims were incorrect (36\%) or  partially correct (46\%): only a small portion were fully correct (18\%).
\section{Discussion and Conclusions}

Large Language Models promise to provide a more rapid and accessible way to performed information seeking tasks compared to search engines: these models can be queried with natural language questions and answers are fluent, summarised and often easy to understand. ChatGPT is currently the most popular LLM with more than 100 million users worldwide. A critical aspect for ensuring these LLM support information seeking tasks to a high standard is that generated answers are correct and are complemented by references that allow users to verify the correctness of the claims. The generation of supporting references is commonly referred to as the process of attribution.

In this paper, we investigated the quality of the attributions of ChatGPT. For this, we instructed ChatGPT to provide evidence in the form of scientific articles, professional magazine articles, Wikipedia pages and other professional website to support the answers it generated to the questions we posed.
Question were taken from a dataset of 160 domain-specific knowledge-based questions. We then performed qualitative assessments of the references that ChatGPT produced for attribution. Our key findings are:

\begin{enumerate}
	\item ChatGPT answers incorrectly about half the time for questions in our domain of focus, Ag-Tech (and in particular crop-growing).
	
	\item ChatGPT does not provide attribution for all answers --- a small portion of answers we obtained did not contain references (9\%), despite they being explicitly required by the prompt we used. This however was often reasonable, as the absence of attribution occurred principally when the model stated it did not know how to answer the questions. In some of the cases where an actual actionable answer was provided but without references, suggestions were made on how to search for possible sources of evidence to corroborate the answer.
	
	\item The majority of the references produced by ChatGPT do not exist (86\%). A concerning aspect of these ``false'' references is that they appear legitimate. For scientific articles, for examples, they would refer to top journals in the field, include volumes and issues that exist, claim the articles are authored by well-known scientists in the field; it would even provide realistic DOIs. Similarly, the name of Wikipedia pages it produces also appear realistic, and their URLs would follow the Wikipedia standard. This is a worrying aspect because users may rely on the fact that these references appear credible and do not proceed to verify whether they exist and support ChatGPT's claims.
	
	\item For the small portion of references produced by ChatGPT that did exist (14\%), we found that to a large extent they did not support the claims the model made about them. We found that this did not depend on whether the answers of ChatGPT were correct either; in fact, for most of the correct answers provided by ChatGPT, the references it gave did not support the answer.
	
\end{enumerate}

These findings provide strong evidence that answers produced by the current version of ChatGPT for information seeking tasks should not be trusted, and that most often than not the fact that the model attributes its answers to references that appear legitimate is not an indication that these resources exist, or that the answer is correct.

\bibliographystyle{ACM-Reference-Format}
\bibliography{bibliography}

\end{document}